\documentclass[lettersize,journal]{IEEEtran}
\usepackage{amsmath,amsfonts}
\usepackage{algorithmic}
\usepackage{array}
\usepackage{textcomp}
\usepackage{stfloats}
\usepackage{url}
\usepackage{verbatim}
\usepackage{graphicx}
\def\BibTeX{{\rm B\kern-.05em{\sc i\kern-.025em b}\kern-.08em
    T\kern-.1667em\lower.7ex\hbox{E}\kern-.125emX}}
\usepackage{balance}

\usepackage{multirow}
\usepackage[table, xcdraw]{xcolor}
\usepackage{subfigure}
\usepackage{algorithm}
\usepackage{listings}

\begin{document}
% \title{Masked Fine-tuning Enables Better Initialization for Token Pruning based Dynamic Vision Transformers}
\title{Bridging The Gaps Between Token Pruning and Full Pre-training via Masked Fine-tuning}
\author{Fengyuan Shi, Limin Wang,~\IEEEmembership{Member,~IEEE}
\thanks{Fengyuan Shi and Limin Wang are with the National Key Labratory for Novel Software Technology, Nanjing University, China.}
\thanks{Email: fengyuanshi1999@gmail.com, lmwang@nju.edu.cn.}}

\markboth{Journal of \LaTeX\ Class Files,~Vol.~18, No.~9, September~2020}%
{How to Use the IEEEtran \LaTeX \ Templates}

\maketitle

\begin{abstract}
Despite the success of transformers on various computer vision tasks, they suffer from excessive memory and computational cost. Some works present dynamic vision transformers to accelerate inference by pruning redundant tokens. A key to improving token pruning is using well-trained models as initialization for faster convergence and better performance. However, current base models usually adopt full image training, i.e., using full images as inputs and keeping the whole feature maps through the forward process, which causes inconsistencies with dynamic models that gradually reduce tokens, including calculation pattern, information amount and token selection strategy inconsistencies. Inspired by MAE which performs masking and reconstruction self-supervised task, we devise masked fine-tuning to bridge the gaps between pre-trained base models used for initialization and token pruning based dynamic vision transformers, by masking image patches and predicting the image class label based on left unmasked patches. Extensive experiments on ImageNet demonstrate that base models via masked fine-tuning gain strong occlusion robustness and ability against information loss. With this better initialization, Dynamic ViT achieves higher accuracies, especially under large token pruning ratios (e.g., 81.9\% vs. 81.3\%, and 62.3\% vs. 58.9\% for DeiT based Dynamic ViT/0.8 and Dynamic ViT/0.3). Moreover, we apply our method into different token pruning based dynamic vision transformers, different pre-trained models and randomly initialized models to demonstrate the generalization ability. 
\end{abstract}


\section{Introduction}
\label{section:introduction}

%遮挡鲁棒性要不要换成对information loss更鲁棒
\IEEEPARstart{V}{ision} transformers have achieved significant successes on many computer vision tasks, such as image classification~\cite{dosovitskiy2020image,touvron2021training,liu2021swin}, object detection~\cite{carion2020end,zhu2020deformable}, semantic segmentation~\cite{zheng2021rethinking,xie2021segformer} action recognition~\cite{arnab2021vivit,bertasius2021space} and action detection~\cite{liu2022end}. However, it is at the cost of huge computations and memory overhead, due to the quadratic time complexity of the number of tokens, making it hard to deploy on resource-constrained devices. In fact, patches in the same image contribute unequally to the final prediction, and there is a large background area leading to redundant computations. Consequently, some works try to reduce spatial redundancy and propose token pruning based dynamic vision transformers, which prune tokens according to attention scores of class token or importance scores predicted by a learnable token selector~\cite{rao2021dynamicvit,pan2021ia,liang2022not,tang2022patch}. 

\begin{figure}[t]
\begin{center}
\includegraphics[width=0.45\textwidth]{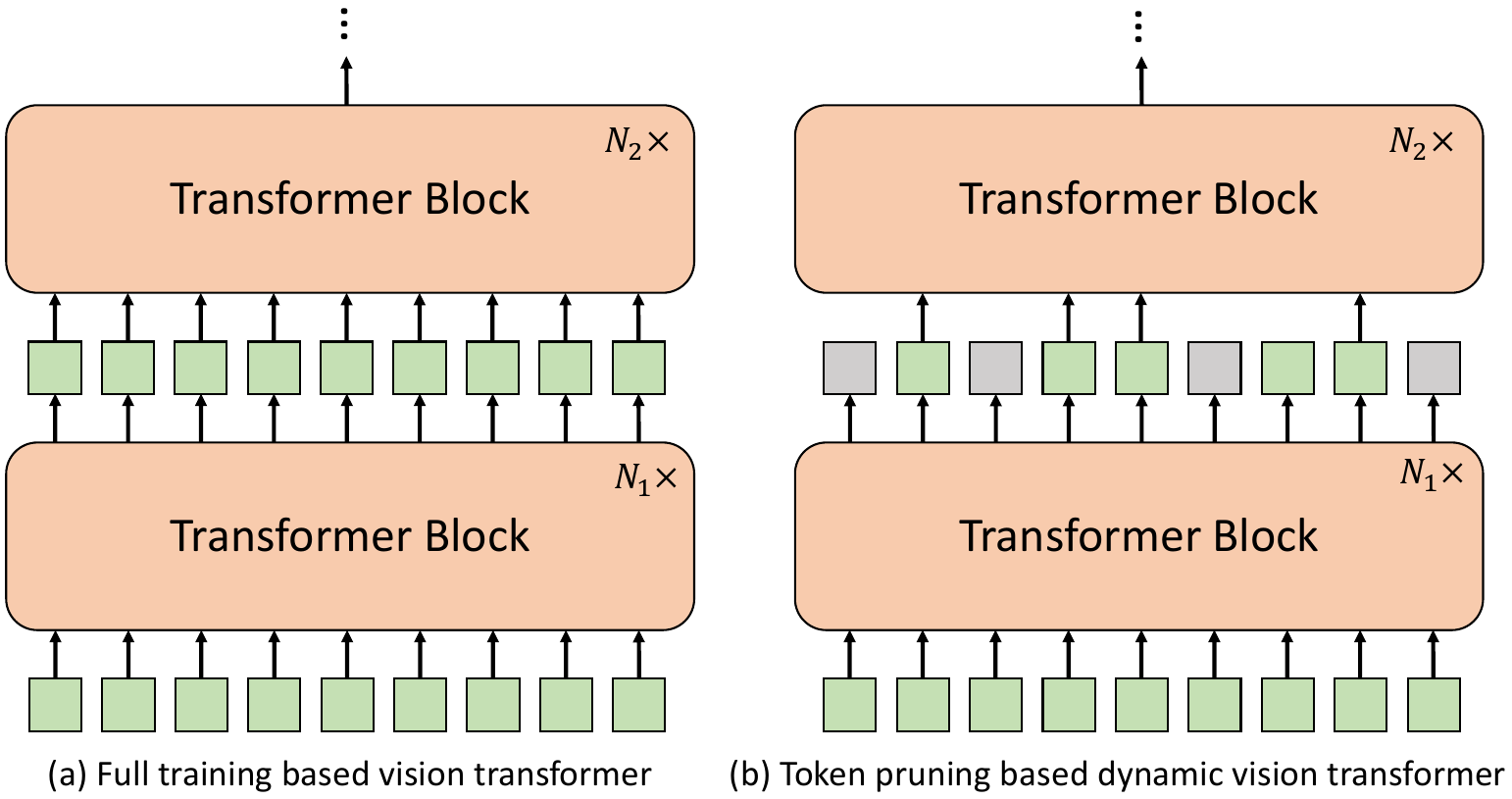}
\end{center}
\caption{Architectures of current base models used for initialization and token pruning based dynamic vision transformers. Full training means using full images as inputs and keeping the whole image structure through the forward computing process. Token pruning keeps the informative tokens and drops redundant tokens during the forward process.}
\label{Introduction:base_and_dynamic_models}
\end{figure}

\begin{figure*}[t]
\centering
\includegraphics[width=1.0\textwidth]{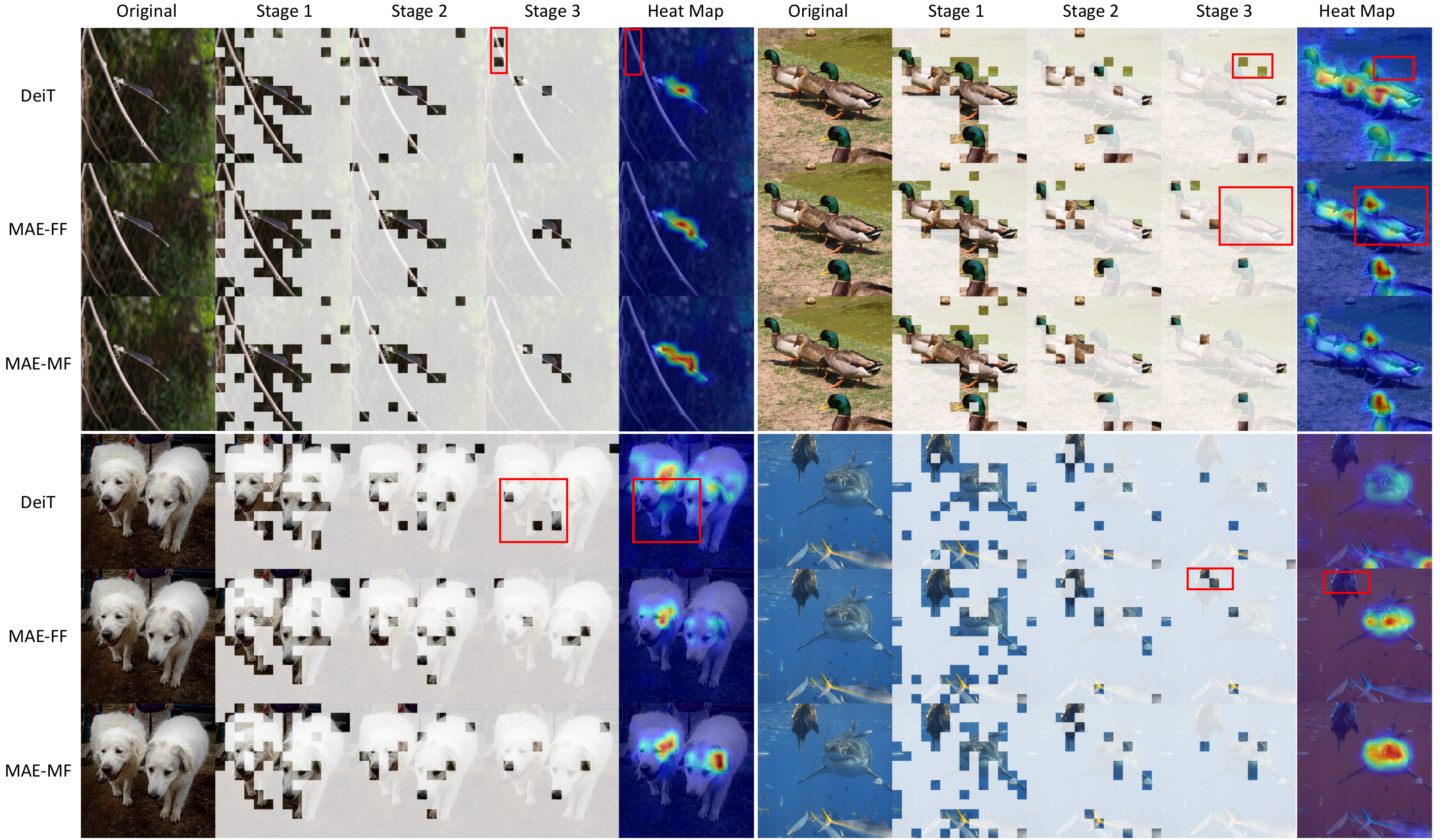}
\caption{Visualization of the token pruning results of Dynamic ViT/0.3 in different pruning stages and the attention heat maps~\cite{chefer2021transformer} of pre-trained base models. MAE-FF means MAE pre-trained model with full fine-tuning, while MAE-MF means MAE pre-trained model with our proposed masked fine-tuning. Attention heat maps of base models with masked fine-tuning are more accurate and focus on the distinguishing parts of the object. Moreover, red boxes show that the final token pruning results of Dynamic ViT/0.3 and attention heat map of base models are significantly different for full training based base models such as DeiT and MAE with full fine-tuning. By masked fine-tuning, Dynamic ViT performs token pruning accurately and keeps informative tokens consistent with that base models focus on.}
\label{Introduction:inconsistency}
\end{figure*}

% 来一张base model和dynamic vit模式差距的图，以及token selection和heat map不一致的问题。
Dynamic vision transformers usually need well-trained base models on ImageNet~\cite{deng2009imagenet} as initialization~\cite{rao2021dynamicvit,pan2021ia,kong2022spvit,liang2022not}, such as DeiT~\cite{touvron2021training}. For one thing, in the early stage of training a dynamic vision transformer from scratch, the token features are not well-encoded. Selecting informative tokens based on these features is unreliable, leading to unstable training and inferior performance. For another, intrinsic sparsity and spatial redundancy exist in the feature map of a well-trained vision transformer, i.e., only a small subset of tokens are useful for the final prediction~\cite{rao2021dynamicvit}, making token pruning more accurate. 
\textbf{However, there are some inconsistencies between the current full pre-training base models like DeiT and token pruning based dynamic vision transformers.} First, the calculation pattern is inconsistent. As shown in Figure \ref{Introduction:base_and_dynamic_models}, current base models usually adopt full training (also called full fine-tuning when fine-tuning pre-trained models such as MAE~\cite{he2022masked}), i.e., using full images as inputs and keeping the whole image structure through the forward process. While token pruning based dynamic vision transformers progressively prune tokens in each stage, and the number of tokens is changing. Second, the amount of information is inconsistent. Dropping patches can be regarded as image occlusions and causes information loss, while there is no loss of information for full training based models due to keeping the whole image structure from beginning to end. Without special training in the case of information loss, the base model has poor ability against occlusion and information loss. Third, the token selection mechanism is inconsistent. For full training based vision transformers, they perform token selection in a soft way without token reduction. They allow wrong selections in shallower layers and continuously refine token selection results to filter out important patches for final prediction in deeper transformer blocks. However, token pruning based dynamic vision transformers select tokens in a hard way, which means that once a token is not selected, it will never be involved in subsequent calculations. Thus the cumulative effect of selection mistakes will be constantly magnified. As shown in Figure \ref{Introduction:inconsistency}, the final token pruning results of Dynamic ViT/0.3 and attention heat map of base models are significantly different with full image training based DeiT and full fine-tuned MAE, which demonstrates the inconsistencies between base models and token pruning based dynamic vision transformers. We argue that {\em these inconsistencies would lead to inferior initialization for token pruning based dynamic vision transformers.}
 % mask input其实就是用一种随机的token selection策略，逼迫模型发现最重要的token

However, previous works all focus on designing better token pruning strategies, but what is a better initialization model is not explored, which is vital for building stronger dynamic vision transformers. To eliminate the inconsistencies and enables stronger token pruning based dynamic vision transformer, we introduce masked fine-tuning, inspired by MAE~\cite{he2022masked} which learns occlusion robustness via masking and reconstruction~\cite{kong2022understanding}. As shown in Figure \ref{Model:figure}, we randomly mask part of the image, and make classification based on unmasked image patches. We also introduce knowledge distillation to further improve the performance of base models. Unlike MAE, we do not reconstruct the masked input patches, but predict the class label based on the masked input image. Masked fine-tuning has some advantages. First, masked fine-tuning uses masked image as input, and the process of dropping patches is similar to dynamic models, mitigating the model inconsistency. Second, masked fine-tuning makes a more complex task, requiring to recognize the class label under the information loss. By masked fine-tuning, base model learns to recognize the most distinctive parts and strengthens ability against occlusion and information loss. Third, random masking introduce different views of the same image, and can be regarded as a kind of regularization and prevent overfitting. 

Like MAE~\cite{he2022masked}, we also study the impact of mask ratio setting. We find that masked fine-tuning with a single lower mask ratio acts as a regularizer and lightly improves model performance, but gains limited occlusion robustness. While masked fine-tuning with a single high mask ratio can improve occlusion robustness but harm the accuracy of base model, due to severe information loss and inconsistent setting between training and inference (masked input vs. full input). To realize a better tradeoff between base model performance and ability against occlusion and information loss, we further introduce masked fine-tuning with hybrid mask ratio. Specifically, we define a mask ratio set including both high mask ratios and low mask ratios, such as [0, 0.25, 0.5, 0.75], and randomly sample mask ratios for the images in the same batch. Compared with single mask ratio, hybrid setting narrows the data distribution between masked training and full-image inference and can better adapt to the token number change in dynamic vision transformers. 

We conduct extensive experiments on ImageNet~\cite{deng2009imagenet}. Experiments demonstrate that via masked fine-tuning, base models are more robust to different occlusion degrees, i.e., different mask ratios, and show significant improvements on large mask ratios. With our better base models as initialization, Dynamic ViT~\cite{rao2021dynamicvit} achieves higher accuracies than their counterparts with full training based models like DeiT, especially for large pruning ratio (e.g., 81.9\% vs. 81.3\%, and 62.3\% vs. 58.9\% for DeiT based Dynamic ViT/0.8 and Dynamic ViT/0.3). We further apply our method into more token pruning based dynamic vision transformers including SPViT~\cite{kong2022spvit}, ATS~\cite{FayyazKJSJSPG22} and EViT~\cite{liang2022not} and achieve consistent performance improvments, which demonstrates the generalization on dynamic model architectures. Moreover, we apply our masked fine-tuning on different pre-trained base models, including masked image modeling based self-supervised pre-trained model MAE~\cite{he2022masked}, contrastive learning based self-supervised pre-trained model MoCo v3~\cite{ChenXH21} and supervised pre-trained model DeiT~\cite{touvron2021training}, to demonstrate the generation on pre-trained models. Meanwhile, we also use our method to train a randomly initialized vision transformer from scratch and further raise the accuracy of dynamic models (73.9\% vs. 58.9\% for DeiT based Dynamic ViT/0.3). To our best knowledge, we are the first work to explore the impact of initialization and introduce masked fine-tuning to bridge the gaps between base models and token pruning based dynamic vision transformers. We hope there are more research works on designing better pre-training strategies for dynamic vision transformers.

\section{Related Work}
\label{section:related_work}
\subsection{Vision Transformers}
Vision Transformers have achieved remarkable success on many computer vision tasks, such as image classification~\cite{dosovitskiy2020image,touvron2021training,liu2021swin}, object detection~\cite{carion2020end,zhu2020deformable}, semantic segmentation~\cite{zheng2021rethinking,xie2021segformer}, action recognition~\cite{arnab2021vivit,bertasius2021space}, point cloud~\cite{zhao2021point,guo2021pct}, low level vison tasks~\cite{liang2021swinir,song2023vision} and so on. ViT~\cite{dosovitskiy2020image} is a pioneering work introducing transformer~\cite{vaswani2017attention} into computer vision, which splits a image into non-overlapped patches and input the token sequence into a pure transformer network. Despite achieving comparable results to state-of-the-art convolutional networks, pre-training on a large-scale dataset is required. DeiT~\cite{touvron2021training} introduces many training techniques to tackle the data-inefficiency problem in ViT. Subsequent works attempt to introduce CNNs~\cite{wu2021cvt,yuan2021incorporating} and hierarchical structures~\cite{liu2021swin,wang2021pyramid} into ViT, and endow ViT with strong priors such as locality and hierarchy, for better adaption to downstream vision tasks.

\subsection{Dynamic Vision Transformers}
Although achieving significant successes, vision transformers suffer from huge computations and memory overhead. Some works propose dynamic vision transformers to reduce spatial redundancy by cutting down the number of tokens. One research line starts from patch size, using fine-grained patch splitting for hard images~\cite{wang2021not} or informative patches~\cite{chen2022coarse}, while corse-grained patch splitting for simple images or uninformative patches. Another research line focuses on token pruning. Dynamic ViT~\cite{rao2021dynamicvit} proposes a lightweight prediction module to estimate the importance score of each token and remove uninformative tokens hierarchically. IA-READ$^2$~\cite{pan2021ia} introduce an interpretable multi-head interpreter to dynamically drop redundant patches. Evo-ViT~\cite{xu2022evo} present a slow-fast token evolution approach, which updates the informative tokens and uninformative tokens with different computation paths for less computations. To reduce information loss, SPViT~\cite{kong2022spvit} and EViT~\cite{liang2022not} aggregate less informative tokens into one package token. PS-ViT~\cite{tang2022patch} discards useless patches in a top-down paradigm, i.e., pruning from the last layer to the first layer, to ensure discriminative patches can be well calculated. And there are some works attempting to prune multiple dimensions, including tokens, heads, blocks and channels~\cite{chen2021chasing,yu2022mia,meng2022adavit,chavan2022vision}.

Current token pruning based dynamic vision transformers all focus on designing a better token selection strategy, but ignore the importance of initialization of dynamic vision transformers. To the best of our knowledge, our paper is the first work to study the role of initialization for dynamic vision transformers, and we propose masked fine-tuning to bridge the gaps between original vision transformers and dynamic vision transformers.

\subsection{Masked Image Modeling}
% （mae introduction, image, video, downstream）
Masked Language Modeling (MLM)~\cite{devlin2018bert} has substantially advanced the development of natural language processing. Recently, researchers have been trying to introduce BERT-style masking and reconstruction task into self-supervised vision transformers~\cite{bao2021beit,he2022masked}. Following BERT~\cite{devlin2018bert}, BEiT~\cite{bao2021beit} propose a masked image modeling task to pre-train vision transformers. Specifically, BEiT tokenizes the original image into visual tokens obtained by the latent codes of discrete VAE~\cite{ramesh2021zero}, randomly replaces some image patches with mask tokens and inputs to the transformer encoder, then recovers the original visual tokens based on the corrupted image patches. Different from BEiT, MAE~\cite{he2022masked} designs an asymmetric encoder-decoder architecture, which only inputs unmasked image patches to the encoder and reconstructs the pixels of masked image patches. SimMIM~\cite{xie2022simmim} proposes a simple framework for masked image modeling, and extends it into hierarchical vision transformers (e.g., Swin Transformer~\cite{liu2021swin}). CAE~\cite{chen2022context} introduces a latent contextual regressor to decouple feature encoding and pretext task, to improve the representation learning capacity. There are also some works trying to realize masked image modeling on hierarchical transformer~\cite{huang2022green, li2022uniform}. The idea of MIM is also extended into other domains, such as video~\cite{tong2022videomae,feichtenhofer2022masked,wang2023videomae}, audio~\cite{huang2022masked} and point cloud~\cite{yu2022point,pang2022masked}.

We also notice that there are some works~\cite{li2022scaling,yang2022attentive} operating on masked images for accelerating the training of CLIP~\cite{radford2021learning}. They use both images and language by introducing language as supervisions, while we only one modality, i.e., images. Moreover, they aim to reduce training time by masking image patches, while we explore the ability against occlusion and information loss brought by masked fine-tuning and aim to provide better initialization for token pruning based dynamic vision transformers.

\section{Masked Fine-tuning}
To bridge the gaps between token pruning based dynamic vision transformers and pre-trained models, we introduce masked fine-tuning, which requires the model to predict the class labels based on masked images. From the respect of calculation processing, randomly dropping the input image patches is similar to the token pruning processing and alleviate the inconsistency of calculation pattern to some extent. Moreover, using masked images as inputs reduces the information amount and alleviate the inconsistency of information loss. By masked fine-tuning, models learn strong ability against occlusion and information loss. As for token selection, randomly masking input image patches adopts a hard token selection strategy similar to token pruning, which means that once a token is not selected, it will never be involved in the subsequent forward process. Therefore, models are forced to  recognize the most distinct parts and strengthen the ability against occlusion and information loss.

\begin{figure}[t]
\centering
\includegraphics[width=0.45\textwidth]{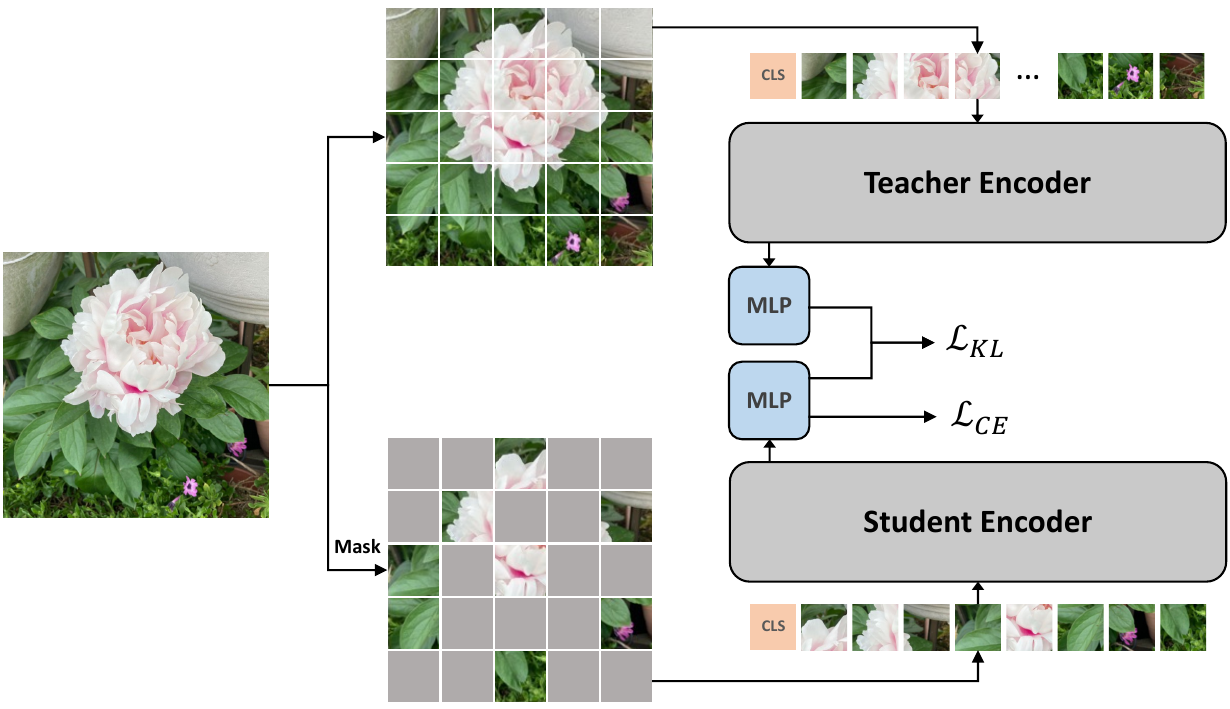}
\caption{An illusion of our masked fine-tuning. We mask some image patches and predict the class label based on unmasked patches for fine-tuning. And we introduce knowledge distillation to further improve the model performance.}
\label{Model:figure}
\end{figure}

\begin{table*}[t]
\begin{center}
\caption{Method comparisons.}
\label{Model:comparison}
% \resizebox{\linewidth}{!}{
\begin{tabular}{c|c|c|c|c|c}
\hline
Method            & Architecture    & Type           & Mask         & Mask Ratio   & Prediction  \\ \hline
MAE               & Encoder-Decoder & Generative     & $\checkmark$ & High         & Pixel      \\ \hline
Full Finetuning   & Encoder         & Discriminative & $\times$     & -            & Class label \\ \hline
Masked Finetuning & Encoder         & Discriminative & $\checkmark$ & High and Low & Class label \\ \hline
\end{tabular}
% }
\end{center}
\end{table*}

\label{section:Method}
\subsection{Training and Inference Procedure}
As shown in Figure \ref{Model:figure}, we first split the image into non-overlapping fixed size patches. Following MAE~\cite{he2022masked}, we randomly sample patches according to the masking strategy, and only input the unmasked patches into the self-supervised pre-trained MAE transformer encoder (can be other per-trained or randomly initialized models) for subsequent training. After a series of transformer blocks, we use the class token of the final block to predict the class label. We calculate the cross entropy loss between the predictions and targets:
\[
    \mathcal{L}_{CE} = \text{CrossEntropy}(\hat{y}, y),
\]
where $\hat{y}$ is the prediction and $y$ is ground truth.

To strengthen the abilities of information reserving and recovering, we use full fine-tuned MAE encoder (or other full fine-tuned encoder) as a teacher and require the prediction of our model to close the teacher's. Specifically, we minimize the Kullback-Leibler divergence between the softmax of logits of the teacher and student model. The distillation loss is as follows:
\[
    \mathcal{L}_{KL} = \text{KL}(\phi(\hat{y} / \tau), \phi(y' / \tau))
\]
where $y'$ is the prediction of the teacher, $\tau$ is the temperature and $\phi$ is softmax function. 

The overall training objective is a combination of these two objectives:
\[
    \mathcal{L} = \mathcal{L}_{CE} + \lambda\mathcal{L}_{KL}
\]
where $\lambda$ is the balance coefficient.

By masked fine-tuning, our vision transformers learn strong occlusion robustness and our models can directly test on any mask ratios during inference. 

\begin{algorithm}[t]
\caption{PyTorch-like Pseudocode of Masked Fine-tuning with Hybrid Mask Ratio}
\definecolor{codeblue}{rgb}{0.25,0.5,0.5}
\definecolor{codekw}{rgb}{0.85, 0.18, 0.50}

\begin{lstlisting}[language=python]
def hybrid_masking(x, hybrid_mask_ratio):
    '''
    x: token sequence (N, L, D)
        N: batch size
        L: number of tokens
        D: feature dimension
    hybrid_mask_ratio: mask ratio set (M, )
        like [0, 0.25, 0.5, 0.75]
    '''
    N, L, D = x.shape

    # sample a mask ratio for each sample in the batch
    keep_ratios = (1 - hybrid_mask_ratio)    # (N,)
    num_keep_tokens = int(L * keep_ratios)    # (N,)
    M = len(hybrid_mask_ratio)
    ratios_choices = randint(0, M, (N,))    # (N, )
    num_sample_tokens = num_keep_tokens[ratios_choices] 
    # (N,)
    
    # produce mask matrix
    noise = rand(N, L)    # (N, L)
    ids_shuffle = argsort(noise, dim=1)    # (N, L)
    ids_restore = argsort(ids_shuffle, dim=1)    # (N, L)

    # 1 keep, 0 remove
    mask = zeros([N, L])    # (N, L)
    for i in range(N):
        mask[i, :num_sample_tokens[i]] = 1
    mask = gather(mask, dim=1, index=ids_restore)
    return mask
\end{lstlisting}
\label{alg:hybrid_masked_fine-tuning}
\end{algorithm}

\subsection{Masking Strategy}
Masking strategy is vital for model performance. We devise two kinds of masking strategies, including single mask ratio and hybrid mask ratio. 

\noindent \textbf{Single mask ratio.}
Like MAE, single mask ratio means using the same mask ratio for all images during training. Specifically, given a mask ratio $\rho$, we randomly sample $\rho N$ tokens under the uniform distribution. As shown in Table \ref{Ablation:masking_strategy}, our masked fine-tuned models show excellent occlusion robustness, i.e., achieving higher accuracy than full fine-tuned models on high mask ratios.

\noindent \textbf{Hybrid mask ratio.}
As shown in Table \ref{Ablation:masking_strategy}, although achieving higher accuracy on large pruning ratios for Dynamic ViT~\cite{rao2021dynamicvit}, masked fine-tuned based models with single mask ratio perform worse. This is caused by the large gap between training and testing data distribution. Therefore, we devise hybrid mask ratio to enable strong ability against occlusion and information loss without sacrificing the performance of the base model. Specifically, we pre-define a mask ratio set including multiple mask ratios ranging from low to high. We set $[0, 0.25, 0.5, 0.75]$ in our experiments. When training, we randomly sample a mask ratio from the mask ratio set for each image in the same batch, thus these images don't share the same mask ratio. Since the model deal with images of different mask ratios during training, it learns stronger occlusion robustness and copes with multiple occlusion situations better. Moreover, dynamic vision transformers usually prune tokens hierarchically and keep different numbers of tokens in different stages. Masked fine-tuning with hybrid mask ratios uses multiple mask ratios during training, thus better adapting to the change of token number in dynamic vision transformers.

Algorithm \ref{alg:hybrid_masked_fine-tuning} provides the pseudo-code of the masking strategy in masked fine-tuning with hybrid mask ratio. We first randomly sample a mask ratio from the mask ratio set for each sample in the batch. Since the mask ratios of different images may be not the same, token numbers of samples in the batch may also be different. For parallelism, we produce a mask matrix to indicate what tokens are masked and unmasked. 

\begin{figure*}[t]
\begin{center}
\subfigure[Top-1 Accuracy under different input mask ratio.]{
\begin{minipage}[t]{0.48\textwidth}
\centering
\includegraphics[width=1.0\textwidth]{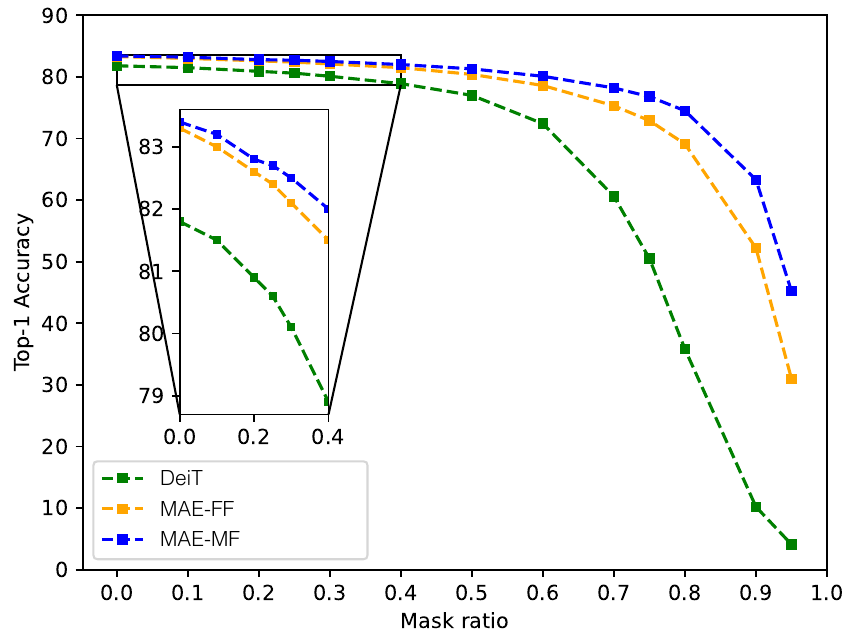}
\end{minipage}
}
\subfigure[Top-1 Accuracy drop under different input mask ratio.]{
\begin{minipage}[t]{0.48\textwidth}
\centering
\includegraphics[width=1.0\textwidth]{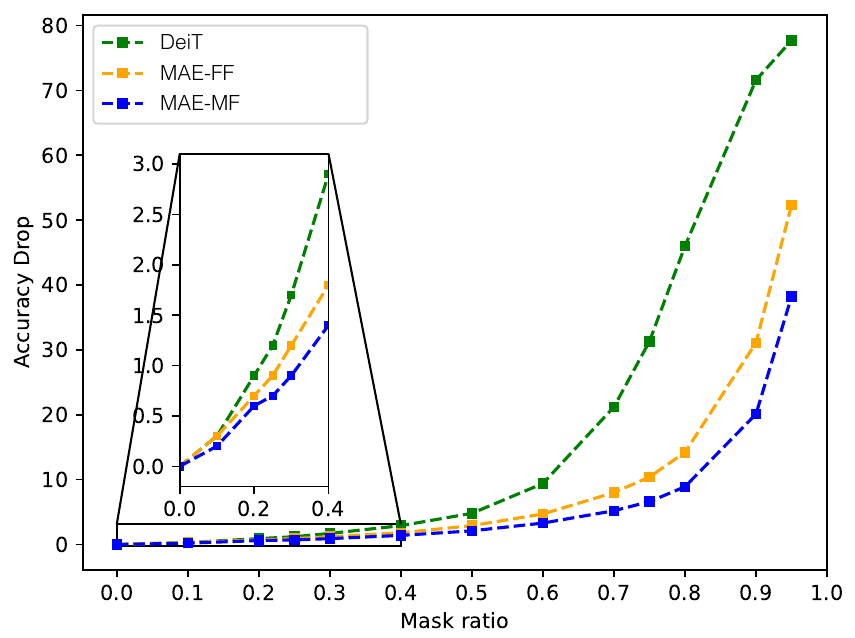}
\end{minipage}
}
\end{center}
\caption{Top-1 Accuracy and accuracy drop of different base models under different mask ratios. MAE-FF means MAE pre-trained model with full fine-tuning, while MAE-MF means MAE pre-trained model with our proposed masked fine-tuning. Masked fine-tuning improves MAE with full fine-tuning and achieves much higher accuracy than DeiT when inputting images with a high mask ratio, which proves its strong occlusion robustness and ability against information loss.}
\label{Ablation:occulsion_curve}
\end{figure*}

\subsection{Discussion With MAE}
Our masked fine-tuning follows MAE~\cite{he2022masked}, randomly samples image patches and only inputs the unmasked patches into the vision transformer. Both methods can improve the ability against occlusion and information loss. 

However, there are some differences between our method and MAE. As shown in Table \ref{Model:comparison}, MAE adopts an encoder-decoder framework and decode the pixels of the masked input patches for reconstruction, while our method predicts the class label based on the masked input image. Differences in prediction targets lead to different semantic levels of learned features.
MAE reconstructs image patch pixels and learns features of low-level semantics~\cite{xie2022simmim,chen2022context}, while masked fine-tuning predicts class labels and learns features of high-level semantics. Low-level visual features are not good at processing high-level understanding tasks like recognition. As shown in Table \ref{Ablation:convergence_table}, using MAE as initialization of Dynamic ViT leads to longer training time and worse performance. Although traditional full fine-tuning can help MAE models learn high-level semantics, the ability against occlusion and information loss will be weakened. However, our masked fine-tuning not only learns high-level semantics for better recognition, but also keeps and strengthens the ability against occlusion and information loss.

Moreover, MAE requires a very high mask ratio (e.g., 75\%) to create a non-trivial self-supervised task. However, such a high mask ratio leads to severe information loss, which is too difficult for image classification and harms the accuracy. Therefore, we carefully design masking strategies by combining low mask ratios and high mask ratios (hybrid mask ratio) to improve model's occlusion robustness for stronger dynamic vision transformers, but not harm the performance of the masked fine-tuned models on ImageNet~\cite{deng2009imagenet}.

\subsection{Initializing Dynamic Vision Transformers}
Our proposed masked fine-tuning is a learning algorithm aiming to improve model occlusion robustness while not harm the performance for full image inputs, and it doesn't introduce any additional parameters and doesn't modify the architecture of the vision transformers. Therefore, our models can be simply used as initialization for token pruning based dynamic vision transformers without any modifications. To validate generalization of our method, we use masked fine-tuned base models as initialization of several popular open source token pruning based dynamic vision transformers. We also apply our method into different pre-trained models or training a randomly initialized model from scratch. Detailed results are in section \ref{section:experiments_generalization}. Using our trained models as initialization, we can build stronger dynamic vision transformers.

\begin{table}[t]
\begin{center}
\caption{Top-1 Accuracy (\%) of different base models on ImageNet under token pruning settings. We adopt Dynamic ViT style hierarchical pruning based on random sampling and cls token attention score, without introducing additional learnable parameters and re-training. Masked fine-tuning shows the best occlusion robustness and ability against information loss.}
\label{Analysis:occlusion_robustness}
% \resizebox{0.7\linewidth}{!}{
\begin{tabular}{c|c|cc}
\hline
\multirow{2}{*}{Model} & \multirow{2}{*}{Full}  & \multicolumn{2}{c}{Token Pruning} \\
                       &                        & random               & attention               \\ \hline
DeiT                             & 81.8                                          & 19.7                 & 26.7                    \\ 
MAE-FF                              & 83.3                                           & 34.4                 & 43.3                    \\ 
MAE-MF           & 83.4                                           & \textbf{44.1}                 & \textbf{51.8}                    \\ \hline
\end{tabular}
% }
\end{center}
\end{table}

\begin{table*}[t]
\begin{center}
\caption{Top-1 Accuracy (\%) of Dynamic ViT initialized by different base models under different token keep ratios in each stage. MAE with masked fine-tuning outperforms DeiT and MAE with full fine-tuning, especially for high keep ratios, demonstrating our masked fine-tuning makes a better initialization and enables stronger dynamic vision transformer.}
\label{Ablation:initialization}
% \resizebox{\linewidth}{!}{
\begin{tabular}{c|c|cccc}
\hline
\multirow{2}{*}{Initialization} & \multirow{2}{*}{Base model} & \multicolumn{4}{c}{Dynamic ViT with different keep ratios}           \\ 
                                 &                      & 0.3          & 0.5         & 0.7         & 0.8                  \\ \hline
DeiT                             & 81.8                 & 58.9 (-22.9) & 78.7 (-3.1) & 81.3 (-0.5) & 81.4 (-0.4)  \\ 
MAE-FF         & 83.3                 & 64.1 (-19.2) & 80.3 (-3.0) & 82.6 (-0.7) & 83.0 (-0.3)  \\ 
MAE-MF       & 83.4                 & 66.1 (-17.3) & 80.8 (-2.6) & 83.0 (-0.4) & 83.3 (-0.1)  \\ \hline
\end{tabular}
\end{center}
\end{table*}

\section{Experiments}
\label{section:experiments}
 In this section, we conduct extensive experiments on ImageNet~\cite{deng2009imagenet} to demonstrate the superiority of masked fine-tuning. In all experiments, we adopt ViT-B as backbone, and choose Dynamic ViT/$\rho$~\cite{rao2021dynamicvit} ($\rho$ is token keep ratio of each stage) to validate the improvements of masked fine-tuning on token pruning based dynamic vision transformers. For masked fine-tuning, we fine-tune MAE self-supervised pre-trained models for 100 epochs following the training techniques used in MAE full fine-tuning. The official code of MAE uses global average pooling on the final layer, while we reproduce the version of using class token to align the setting with DeiT and Dynamic ViT. After masked fine-tuning, we use it to initialize Dynamic ViT and train for 30 epochs following Dynamic ViT. We set [0, 0.25, 0.5, 0.75] as default hybrid mask ratio and use soft distillation as default distillation method. The temperature $\tau$ and loss balance coefficient $\lambda$ are set to 1.0. 

For simplicity, we denote MAE self-supervised pre-train as `MAE', MAE self-supervised pre-train with full fine-tuning as `MAE-FF', and MAE self-supervised pre-train with our proposed masked fine-tuning as `MAE-MF'.

\subsection{Main Results}
\noindent \textbf{Stronger ability against occlusion and information loss of base models.}
Masked fine-tuning forces the model to predict the class label of obscured images in the presence of information loss, endowing the base model with occlusion robustness and strong ability against information loss, thus enabling stronger dynamic vision transformers. We perform an in-depth experimental investigation into this nature of our masked fine-tuning, from two experimental settings including using masked images as input and Dynamic ViT style token pruning on ImageNet. 

First, we explore the performance of base model used for initialization under different occlusion levels by controlling the mask ratio of input images, to demonstrate masked fine-tuning is more robust than full image training, including full training DeiT and MAE with full fine-tuning. As shown in Figure \ref{Ablation:occulsion_curve}, masked fine-tuning achieves best performance on all mask ratios, and shows significant advantages under extreme occlusion and information loss. Specifically, the accuracy of DeiT starts plunging when mask ratio increases to 0.5, while the accuracy of masked fine-tuning decreases clearly when mask ratio rises to 0.7. Surprisingly, even with an extreme mask ratio like 0.95, masked fine-tuning gains 45.2 \% accuracy, while DeiT only gets 4.1\%. MAE-FF performs better than DeiT since the mask and reconstruction self-supervised pre-train task, but subsequent full fine-tuning will harm the occlusion robustness. Instead, our proposed masked fine-tuning mask input images and predict the class label during fine-tuning, and achieves better performance. 

For token pruning setting, we keep 70\% left tokens in each stage, which is the same as Dynamic ViT. Different from the learnable token selection module, we simply use random sampling and cls token's attention for token pruning. It is worth noting that we do not introduce any additional learnable parameters and directly test the base models. The results are shown in Table \ref{Analysis:occlusion_robustness}. From the table, we can find that full training based DeiT performs worst, e.g., 60.1 percent drop for random token pruning. MAE with full fine-tuning performs better which benefits from mask and reconstruct self-supervised pre-train task, but still significantly lagging behind masked fine-tuning. These phenomena suggest that full-image training will corrupt the model ability against occlusion and information loss, and is sub-optimal as initialization for dynamic vision transformer. 

\noindent \textbf{Enabling stronger token pruning based dynamic vision transformers.} We further test Dynamic ViT with different base models as initialization under different token keep ratio in Table \ref{Ablation:initialization}. From the table, we can see that our method achieves least accuracy drops against the base models on all keep ratios. For a mask ratio of 0.8, our Dynamic ViT is only 0.1 lower than its base model. Surprisingly, when using smaller keep ratios and pruning more tokens, Dynamic ViT with masked fine-tuning shows significant advantages within all methods, i.e., the smallest accuracy drops against their base models. The process of masking input and predicting class labels in masked fine-tuning eliminates inconsistencies between base models and dynamic vision transformers, and endows models with strong ability against occlusion and information loss, thus enabling stronger token pruning based dynamic vision transformers.

\begin{table}[t]
\begin{center}
\caption{Top-1 accuracy (\%) of Dynamic ViT/0.7 training from scratch and initialized with different base models on ImageNet. With well-trained models as initialization, Dynamic ViT converges faster and gets better accuracy. Under the same training epochs, Dynamic ViT with masked fine-tuning outperforms counterparts with other base models as initialization.}
\label{Analysis:use_initialization}
% 证明Initialization对Dynamic vision transformer的重要性
% \resizebox{\linewidth}{!}{
\begin{tabular}{c|c|c|c}
\hline
Initialization      & Epochs & Base model & Dynamic ViT/0.7 \\ \hline
No                  & 30    & - & 16.3           \\ 
No                  & 300   & - & 56.9           \\ \hline
DeiT                & 30    & 81.8 & 81.3           \\ 
MAE                 & 30    & - & 68.6           \\
MAE                 & 100   & - & 75.2            \\
MAE-FF              & 30    & 83.3 & 82.6           \\ 
MAE-MF              & 30    & 83.4 & 83.0           \\ \hline
\end{tabular}
% }
\end{center}
\end{table}

\begin{figure}[t]
\begin{center}
\includegraphics[width=0.45\textwidth]{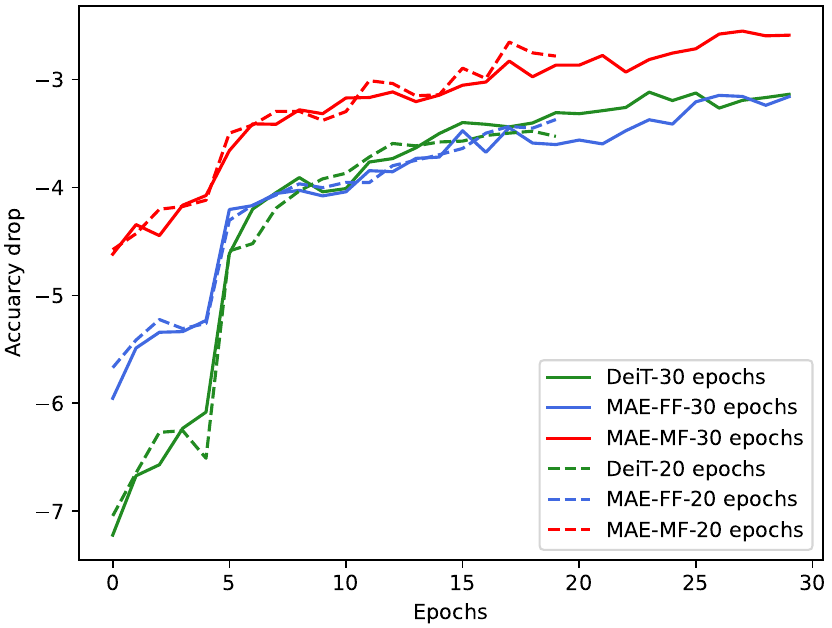}
\end{center}
\caption{Convergence curves of Dynamic ViT/0.5 with different base models as initialization. Dynamic ViT with masked fine-tuning quickly narrows the accuracy gap with base models from the very beginning of training, proving that masked fine-tuning eliminates inconsistencies between base models for initialization and token pruning based dynamic vision transformers.}
\label{Ablation:convergence_curve}
\end{figure}

\begin{table}[t]
\begin{center}
\caption{Top-1 Accuracy of Dynamic ViT/$\rho$ initialized by different base models under different training time. $\rho$ is keep ratio. Dynamic ViT with full training based base modes like DeiT and full fine-tuned MAE as initialization incurs a more obvious accuracy drop for lower keep ratios when reducing training time.}
\label{Ablation:convergence_table}
\begin{tabular}{c|c|ccc}
\hline
\multirow{2}{*}{Initialization} & \multirow{2}{*}{Epochs} & \multicolumn{3}{c}{Dynamic ViT/\textbf{$\rho$}} \\
                                &                         & 0.3              & 0.5              & 0.7              \\ \hline
DeiT                            & 30                      & 58.9             & 78.7             & 81.3             \\
DeiT                            & 20                      & 56.7             & 78.3             & 81.3             \\ \hline
MAE-FF          & 30                      & 64.1             & 80.3             & 82.6             \\
MAE-FF         & 20                      & 62.3             &   80.0           &     82.5         \\ \hline
MAE-MF          & 30                      & 66.1             & 80.8             & 83.0             \\
MAE-MF          & 20                      & 65.0             & 80.6             & 82.9             \\ \hline
\end{tabular}
\end{center}
\end{table}

\begin{figure*}[t]
\begin{center}
\includegraphics[width=0.78\textwidth]{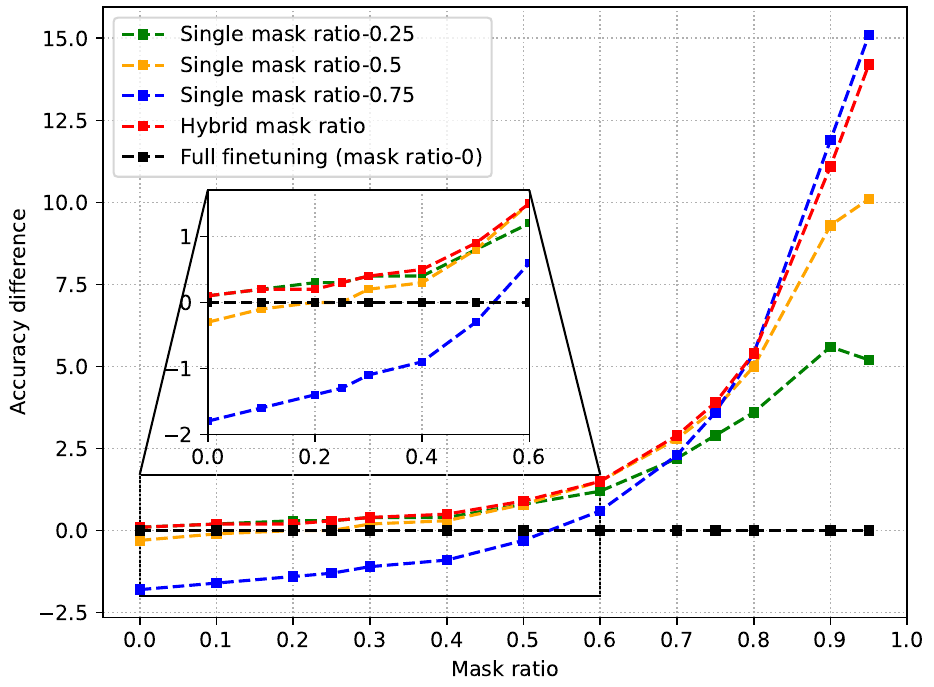}
\end{center}
\caption{Accuracy difference of masked fine-tuned MAE with different masking strategies compared to MAE with full fine-tuning under different input mask ratio. A higher mask ratio significantly improves Dynamic ViT on small keep ratios, but harms the performance of base models. Hybrid mask ratio combines high and low mask ratios for better tradeoff.}
\label{Occulsion:masking_strategies}
\end{figure*}

\noindent \textbf{Faster Convergence.}
We first compare Dynamic ViT training from scratch and initialized by base models. From table \ref{Analysis:use_initialization}, we can find that well-trained models as initialization enables Dynamic ViT with much faster convergence and much better performance and our masked fine-tuning performs best. With only 30 epochs, Dynamic ViT with initialization achieves slightly poorer accuracy than their base models, while only 16.3\% top-1 accuracy for models training from scratch. Although increasing the training time such as 300 epochs like DeiT improves the accuracy of Dynamic ViT without initialization, the performance is still far behind. By comparing Dynamic ViT with self-supervised pre-trained MAE and fine-tuned MAE, we can find that features with high level of semantics is essential for improving token pruning based dynamic vision transformers. 

We further compare Dynamic ViT initialized by different base models with varying keep ratios under different training time, and the results are shown in Table \ref{Ablation:convergence_table}. Reducing the training epochs from 30 to 20, Dynamic ViT with DeiT as base model yields a higher accuracy drop than that with our masked fine-tuned models, such as 0.4\% and 0.2\% for Dynamic ViT/0.5. For a smaller keep ratio 0.3, DeiT drops more (2.2\% vs. 1.1\%), which indicates that full image training based DeiT as base models exists a bigger gap with token pruning based dynamic vision transformers than masked fine-tuned based base models. We also plot the convergence curves in Figure \ref{Ablation:convergence_curve}. Dynamic ViT with masked fine-tuned base models narrows the accuracy gap with base model quickly from the first epoch. Moreover, the accuracy drop curve of Dynamic ViT training 20 epochs is almost above that of training 30 epochs for masked fine-tuning, while below for DeiT. 

Masked fine-tuning uses masked image as input, and the process of dropping patches is similar to dynamic models, which mitigates the model inconsistency between base models used for initialization and dynamic vision transformers. Moreover, dynamic vision transformers usually prune tokens hierarchically and keep different number of tokens in different stages. Masked fine-tuning with hybrid mask ratios uses multiple mask ratios and sample different number of patches for the images in the same bath during training, thus better adapts to the change of token number in dynamic vision transformers.

\subsection{Ablation Study}
\noindent \textbf{Masking Strategies.}
Masking is vital for masked fine-tuning. A naive mask strategy is randomly sampling a single fixed ratio of image patches such as 25\% and drop masked patches, just like MAE. We explore the impact of mask ratio. From Table \ref{Ablation:masking_strategy}, we can find that as mask ratio increases, the performance of base model decreases while initialized Dynamic ViT reduces the accuracy drop against the base model, especially when keeping a small number of tokens. Specifically, via masked fine-tuning with a 0.25 mask ratio, the accuracy of MAE is slightly higher than the full fine-tuning counterpart of MAE (83.4\% vs. 83.3\%) and the accuracy gap closes, e.g., -0.3\% vs.-0.7\% for Dynamic ViT/0.7. For mask fine-tuning with a large mask ratio of 0.75, the accuracy of base model drops significantly (81.5\% vs. 83.3\%), but the corresponding Dynamic ViT obtains minimal accuracy drop against base model, even achieving higher accuracy than base model for Dynamic ViT/$\rho$ (82.0\% vs. 81.5\%). 

This phenomenon is different from the conclusion of MAE which requires a large mask ratio. MAE inputs unmasked image patches into the encoder and reconstruct pixels of masked patches. A large mask ratio can reduce spatial redundancy and avoid the masked patch being simply recovered from neighboring patches without high-level understanding. But such a high mask ratio leads to severe information loss, which is too difficult for image classification and harms the accuracy. 

To improve the base model's occlusion robustness without sacrificing the base model's performance, we introduce masked fine-tuning with hybrid mask ratio, i.e., allowing different mask ratios for images in the same batch during training. As shown in Table \ref{Ablation:masking_strategy}, masked fine-tuning with hybrid mask ratio realizes better trade-off and bridges the gap between base models for initialization and dynamic vision transformers.

We also test base models with masked fine-tuning of different mask strategies on masked input images. We subtract the accuracy of MAE with full fine-tuning from the accuracies of MAE masked fine-tuned with different masking strategies, and plot the accuracy difference under different mask ratios in Figure \ref{Occulsion:masking_strategies}. For masked fine-tuning with single mask ratios, the higher the mask ratio, the higher the accuracy of base models on images with high mask ratios, while the lower the accuracy on unmasked images and images with low mask ratios. Comparing masked fine-tuning of different single mask ratios with full fine-tuning, we can find a low mask ratio, such as 0.25, lightly improves the performance of base model for unmasked images (83.4\% vs. 83.3\%), but gains limited occlusion robustness (least accuracy improvements among all masking strategies). On the contrary, a high mask ratio significantly improves the ability against occlusion and information loss. For example, masked fine-tuning with single mask ratio of 0.25 has 15.1\% percent increase compared to full fine-tuning for 95\% masked input images, but 1.8\% percent decrease for unmasked images. 

To improve ability against occlusion and information loss without sacrificing performance of base models for unmasked images, we introduce masked fine-tuning with hybrid mask ratios combining both low and high mask ratios. Figure \ref{Occulsion:masking_strategies} shows masked fine-tuning with hybrid mask ratios achieves best performance on almost all input mask ratios.

\noindent \textbf{Hybrid mask ratio setting.}
We further study the hybrid mask ratio setting, including low range, high range, both range with dense interval, and both range with sparse interval. The detailed designs are shown in Table \ref{Ablation:hybrid_mask_ratio}. Similar to masked fine-tuning with single mask ratio, we also find that increasing mask ratio will harm the performance of base models, but improve Dynamic ViT. An interesting finding is that MAE with low-range hybrid masked fine-tuning outperforms its full fine-tuning counterpart, i.e., 0.4\% improvements. It indicates our hybrid masked fine-tuning plays a certain role of regularization and can alleviate the overfitting problem. But too high mask ratio causes severe information loss and thus impairs the performance of base models. Combining low and high range of mask ratios is a better solution to balance performance of base models and dynamic vision transformers. We also compare dense interval and sparse interval. Table \ref{Ablation:hybrid_mask_ratio} shows hybrid masked fine-tuning using both low range and high range with sparse interval is better and we take this as our default setting.

\noindent \textbf{Distillation methods.}
Following DeiT, we adopt soft distillation and hard distillation. Soft distillation minimizes the Kullback-Leibler divergence between the softmax logits of the teacher and student model. Hard distillation uses the hard decision of the teacher as a true label to compute cross entropy. The results are shown in Table \ref{Ablation:distillation}. We can see that soft distillation improves the performance of both pre-trained models used for initialization and Dynamic ViT, and we choose soft distillation as the default setting.

\begin{table*}[t]
\begin{center}
\caption{Ablation for masking strategies. A higher mask ratio significantly improves Dynamic ViT on small keep ratios, but harms the performance of base models. Hybrid mask ratio combines high and low mask ratios for better tradeoff. Default settings are marked in gray.}
\label{Ablation:masking_strategy}
\begin{tabular}{c|c|cccc}
\hline
\multirow{2}{*}{Masking Strategies}                       & \multirow{2}{*}{Mask Ratio} & \multicolumn{4}{c}{Top-1 Accuracy (\%)}                                                  \\
                                                          &                             & Base model     & Dynamic ViT/0.3 & Dynamic ViT/0.5 & Dynamic ViT/0.7\\ \hline
No                                                        & 0                           &
83.3      & 64.1 (-19.2)           & 80.3 (-3.0)            & 82.6 (-0.7)                               \\ 
Single Mask Ratio                                         & 0.25                        & 83.4      & 65.7 (-17.7)           & 80.7 (-2.7)           & 83.0 (-0.4)                           \\
Single Mask Ratio                                         & 0.5                         & 83.0      & 65.6  (-17.4)          & 80.9 (-2.1)           & 82.9 (-0.1)                           \\
Single Mask Ratio                                         & 0.75                        & 81.5        & 66.4 (-15.1)         & 80.1 (-1.4)           & 82.0 (+0.5)                           \\
\rowcolor[HTML]{E7E6E6} 
Hybrid Mask Ratio                                         & {[}0, 0.25, 0.5, 0.75{]}    & 83.4      & 66.1 (-17.3)           & 80.8 (-2.6)           & 83.0 (-0.4)                      \\ \hline
\end{tabular}
\end{center}
\end{table*}

\begin{table*}[t]
\begin{center}
\caption{Ablation for hybrid mask ratio settings. Mask fine-tuning with low range of mask ratios acts as a regularizer and improves the performance of base models, but gains limited improvements for Dynamic ViT. Combing low and high ranges of mask ratios realizes a better tradeoff between the performance of base model and Dynamic ViT. Default settings are marked in gray.}
\label{Ablation:hybrid_mask_ratio}
\begin{tabular}{c|c|ccc}
\hline
\multirow{2}{*}{Strategy} & \multirow{2}{*}{Mask Ratio}                                              & \multicolumn{3}{c}{Top-1 Accuracy (\%)} \\
                          &                                                                          & Base model  & Dynamic ViT/0.7 & $\Delta$ \\ \hline
No                        & 0                                                                        & 83.3     & 82.6            & -0.7  \\
Low                       & {[}0, 0.1, 0.2, 0.25, 0.3, 0.4, 0.5{]}                                   & 83.7     & 83.1            & -0.6  \\
High                      & {[}0.5, 0.6, 0.7, 0.75, 0.8, 0.9, 1.0{]}                                 & 81.6     & 81.6            & 0     \\
Both (Dense)              & {[}0, 0.1, 0.2, 0.25, 0.3, 0.4, 0.5, 0.6, 0.7,   0.75, 0.8, 0.9, 0.95{]} & 82.9     & 82.6            & -0.3  \\
\rowcolor[HTML]{E7E6E6} 
Both (Sparse)     & {[}0, 0.25, 0.5, 0.75{]}                                                 & 83.4     & 83.0            & -0.4  \\ \hline
\end{tabular}
\end{center}
\end{table*}

\begin{table}[t]
\begin{center}
\caption{Ablation for distillation methods. Default settings are marked in gray.}
\label{Ablation:distillation}
\begin{tabular}{c|ccc}
\hline
Distillation & Base model  & Dynamic ViT/0.7 & $\Delta$ \\ \hline
No                            & 82.9     & 82.4            & -0.5  \\
Hard                          & 82.7     & 82.0            & -0.7  \\
\rowcolor[HTML]{E7E6E6} 
Soft                          & 83.4     & 83.0            & -0.4  \\ \hline
\end{tabular}
\end{center}
\end{table}

\subsection{Generalization Ability}
\label{section:experiments_generalization}

\noindent \textbf{Generalization on different base models.} 
In addition to masked image modeling based MAE~\cite{he2022masked} self-supervised pre-trained models, we also apply our masked fine-tuning on other pre-trained models, including contrastive learning based self-supervised pre-trained model MoCo v3~\cite{ChenXH21} and supervised pre-trained model DeiT~\cite{touvron2021training}. For MoCo v3, we adopt the same training tricks and schedule as full fine-tuning, but with our masked fine-tuning. For DeiT, we continue training well-trained DeiT for another 30 epochs via masked fine-tuning, following the training settings of fine-tuning DeiT on 384 resolution images. Table \ref{Generalization:more_base_models} shows that Dynamic ViT with masked fine-tuned MoCo v3 significantly outperforms its counterpart with full fine-tuned MoCo v3, such as 3.8 lower accuracy drop against the base model for Dynamic ViT/0.3 (-14.6 vs. -18.4). After masked fine-tuning for another 30 epochs, the well-trained DeiT not only provides a better initialization for token pruning based dynamic vision transformers, but also improves its performance.

We also train a randomly initialized DeiT from scratch with our masked fine-tuning, adopting the same training strategies as full training based DeiT. Although the accuracy of base model is lower (81.3 vs. 81.8), the corresponding Dynamic ViT shows amazing improvements such as only 7.4 accuracy drop for Dynamic ViT/0.3 and 1.1 accuracy drop for Dynamic ViT.0.5. 

Masked fine-tuning achieves consistent improvements on supervised pre-trained models, constrastive learning based self-supervised pre-trained models and masked image modeling based self-supervised pre-trained models, which demonstrate the good model generalization of our proposed masked fine-tuning. Without pre-training and training from scratch, our masked fine-tuning shows much greater strengths on improving token pruning based dynamic vision transformers.

\noindent \textbf{Generalization on different token pruning based dynamic vision transformers.} 
Our proposed masked fine-tuning is a learning algorithm that does not introduce any additional learnable parameters and does not modify the architecture of the vision transformers. Therefore, we can simply load weights of our masked fine-tuned models to token pruning based dynamic vision transformers and adopt the same follow-up training without any modifications.

To validate the generalization of masked fine-tuning, we apply masked fine-tuning into several popular open source token pruning based dynamic vision transformers Dynamic ViT~\cite{rao2021dynamicvit}, SPViT~\cite{kong2022spvit}, ATS~\cite{FayyazKJSJSPG22} and EViT~\cite{liang2022not}. The results in Table \ref{Generalization:more_token_pruning_based_dynamic_vision_transformers} show the consistent improvements of our masked fine-tuning on these dynamic vision transformers.

\begin{table*}[t]
\begin{center}
\caption{Generalization of masked fine-tuning on different base models. `FF' means full fine-tuning and `MF' means masked fine-tuning. We denote DeiT training from scratch with our masked fine-tuning as `DeiT-MF$^\ast$'. Masked fine-tuning works for various kinds of pre-trained models, such as supervised pre-trained model DeiT, constrastive learning based self-supervised pre-trained model MoCo v3 and masked image modeling based self-supervised pre-trained model MAE. By applying masked fine-tuning from scratch, Dynamic ViT achieves the least accuracy drops against its base model.}
\resizebox{\linewidth}{!}{
\begin{tabular}{c|c|c|c|c|c|c}
\hline
Initialization & Type                                                                                             & Base model & Dynamic ViT/0.3 & Dynamic ViT/0.5 & Dynamic ViT/0.7 & Dynamic ViT/0.8 \\ \hline
DeiT           & \multirow{3}{*}{Supervised}                                                                      & 81.8       & 58.9 (-22.9)    & 78.7 (-3.1)     & 81.3 (-0.5)     & 81.3 (-0.4)     \\
DeiT-MF        &                                                                                                  & 82.0       & \underline{62.3 (-19.7)}    & \underline{79.1 (-2.8)}     & \underline{81.5 (-0.5)}     & \underline{81.9 (-0.1)}     \\
DeiT-MF$^\ast$      &                                                                                                  & 81.3       & \textbf{73.9 (-7.4)}     & \textbf{80.2 (-1.1)}     & \textbf{81.1 (-0.2)}     &  \textbf{81.3 (-0.0)}               \\ \hline
MoCo v3-FF     & \multirow{2}{*}{\begin{tabular}[c]{@{}c@{}}Self-supervised\\ (contrastive learning)\end{tabular}}  & 83.0       & 64.6 (-18.4)    & 80.3 (-2.7)     & 82.3 (-0.7)     & 82.6 (-0.4)     \\
MoCo v3-MF     &                                                                                                  & 83.1       & \textbf{68.5 (-14.6)}    & \textbf{80.8 (-2.3)}     & \textbf{82.6 (-0.5)}     & \textbf{83.0 (-0.1)}     \\ \hline
MAE-FF         & \multirow{2}{*}{\begin{tabular}[c]{@{}c@{}}Self-supervised\\ (masked image modeling)\end{tabular}} & 83.3       & 64.1 (-19.2)    & 80.3 (-3.0)     & 82.6 (-0.7)     & 83.0 (-0.3)     \\
MAE-MF         &                                                                                                  & 83.4       & \textbf{66.1 (-17.3)}    & \textbf{80.8 (-2.6)}     & \textbf{83.0 (-0.4)}     & \textbf{83.3 (-0.1)}     \\ \hline
\end{tabular}
}
\end{center}
\label{Generalization:more_base_models}
\end{table*}

\begin{table*}[t]
\begin{center}
\caption{Generalization of masked fine-tuning on different token pruning based dynamic vision transformers.}
\begin{tabular}{c|c|c|c|c|c}
\hline
Initialization & Base model & Dynamic ViT/0.7 & SPViT/0.7   & ATS         & EViT/0.7    \\ \hline
DeiT           & 81.8       & 81.3 (-0.5)     & 80.9 (-0.9) & 81.3 (-0.5) & 80.5 (-1.3) \\
MAE-FF         & 83.3       & 82.6 (-0.7)     & 82.5 (-0.8) & 82.8 (-0.5) & 82.4 (-0.9) \\
MAE-MF         & 83.4       & \textbf{83.0 (-0.4)}     & \textbf{82.8 (-0.6)} & \textbf{83.0 (-0.4)} & \textbf{82.7 (-0.7)} \\ \hline
\end{tabular}
\end{center}
\label{Generalization:more_token_pruning_based_dynamic_vision_transformers}
\end{table*}

\begin{table}[t]
\begin{center}
\caption{Comparison with state-of-the-art token pruning based dynamic vision transformers. The models in the upper part are based on DeiT, and the models in the lower part are based on MAE. "MF" means dynamic models with our masked fine-tuned models as initialization.}
\label{Experiments:SOTA}
\begin{tabular}{l|c|c}
\hline
Method                   & Top-1 Acc (\%) & FLOPs (G) \\ \hline
\rowcolor[HTML]{E7E6E6} 
DeiT~\cite{touvron2021training}                   & 81.8           & 17.6      \\
IA-READ$^2$~\cite{pan2021ia}  & 80.9   & 11.8      \\
SPViT/0.7~\cite{kong2022spvit}                  & 80.9    & 11.6      \\
Evo-ViT~\cite{xu2022evo}                    & 81.3    & -         \\
PS-ViT~\cite{tang2022patch}                     & 81.5    & \textbf{9.8}       \\ 
Dynamic ViT/0.7~\cite{rao2021dynamicvit}            & 81.3    & 11.4      \\
Dynamic ViT/0.8~\cite{rao2021dynamicvit}             & 81.4    & 13.3      \\
Dynamic ViT/0.7-MF~\cite{rao2021dynamicvit} (ours)            & 81.5    & 11.4 \\
Dynamic ViT/0.8-MF~\cite{rao2021dynamicvit} (ours)            & \textbf{81.9}    & 13.3 \\
\hline
\rowcolor[HTML]{E7E6E6} 
MAE~\cite{he2022masked}                   & 83.4          & 17.6      \\
SPViT/0.7~\cite{kong2022spvit}                    & 82.5    & 11.6      \\
Dynamic ViT/0.7~\cite{rao2021dynamicvit}             & 82.6    & \textbf{11.4}      \\
Dynamic ViT/0.8~\cite{rao2021dynamicvit}             & 83.0    & 13.3      \\
SPViT/0.7-MF~\cite{kong2022spvit} (ours)                    & 82.8    & 11.6      \\
Dynamic ViT/0.7-MF~\cite{rao2021dynamicvit} (ours)             & 83.0    & \textbf{11.4}      \\
Dynamic ViT/0.8-MF~\cite{rao2021dynamicvit} (ours)             & \textbf{83.3}    & 13.3      \\ \hline
\end{tabular}
% }
\end{center}
\end{table}

\subsection{Comparisons with SOTA}
In table \ref{Experiments:SOTA}, we compare dynamic vision transformer with our masked fine-tuned base models as initialization and other token pruning based dynamic vision transformers with full-trained base models as initialization. For both DeiT-based and MAE-based parts, Dynamic ViT/0.8 with masked fine-tuning achieves the highest accuracies. And Dynamic ViT/0.7 with masked fine-tuning achieves the least accuracy drop other than PS-ViT under similar computation overhead. For the same dynamic vision transformers, dynamic models with our proposed masked fine-tuning outperforms their counterparts with full-trained base models, such as 81.4\% vs. 81.9\% for Dynamic ViT/0.8 (DeiT) and 82.5\% vs. 82.8\% for SPViT/0.7 (MAE). Surprisingly, there is only 0.1\% accuracy drop for Dynamic ViT/0.8 (MAE) with masked fine-tuning, and even 0.1\% accuracy rise for Dynamic ViT/0.8 (DeiT), but FLOPs reduce from 17.6G to 13.4G.

\begin{figure*}[t]
\begin{center}
\includegraphics[width=0.85\textwidth]{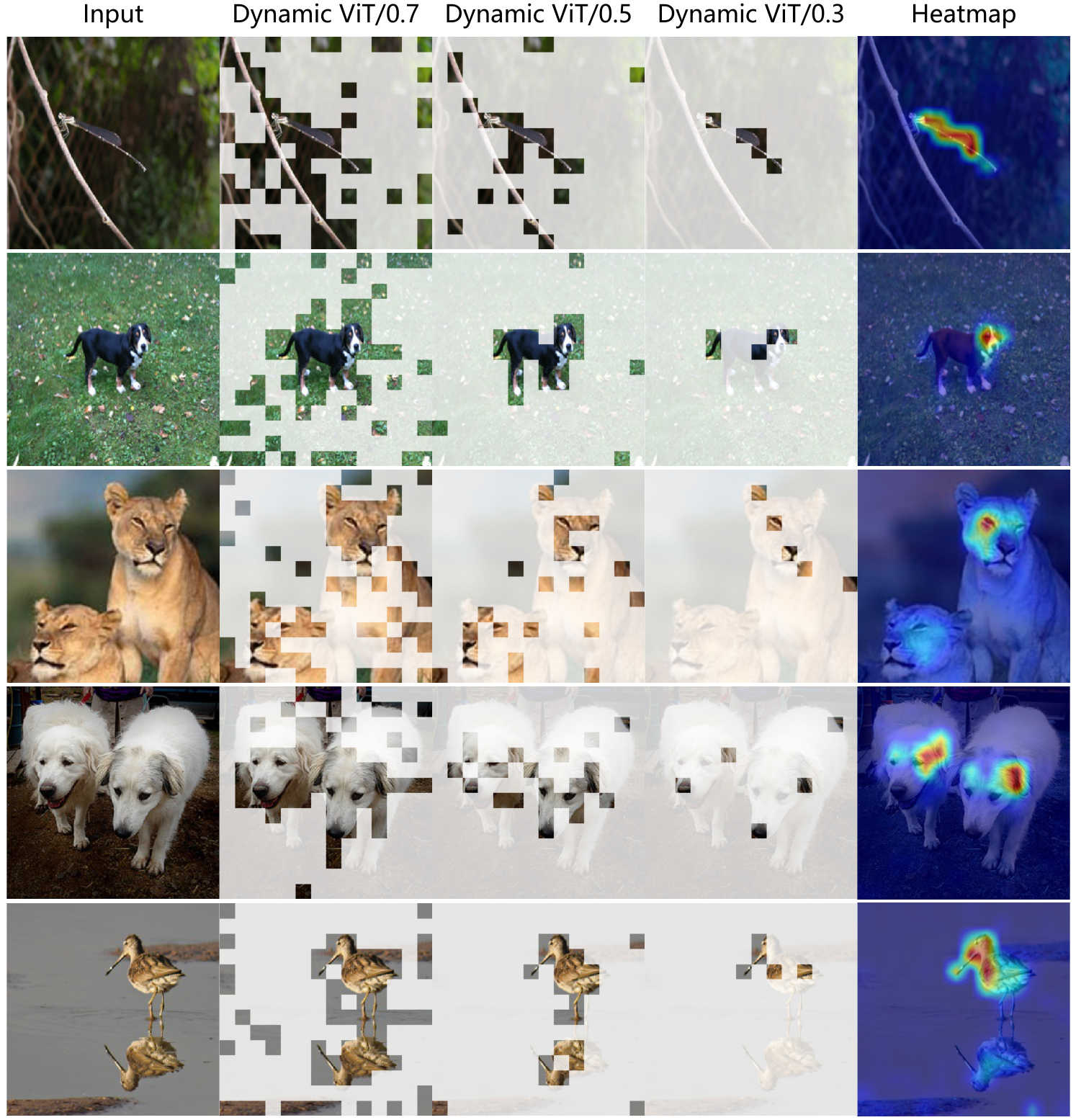}
\end{center}
\caption{Visualization of the results of token pruning produced by Dynamic ViT with different keep ratios and the attention heatmaps~\cite{chefer2021transformer} of base model trained with masked fine-tuning. Vision transformers fine-tuned with our masked fine-tuning and corresponding Dynamic ViT can focus on the fore-ground area and informative parts accurately. }
\label{Visualization}
\end{figure*}

\subsection{Visualization}
We visualize the token pruning results of final stage of Dynamic ViT with different keep ratios, which is initialized by our masked fine-tuned base models. From Figure \ref{Visualization}, we can see that most kept image patches are concentrated in the foreground area, which demonstrates that Dynamic ViT with masked fine-tuning as initialization can effectively filter out informative fore-ground image patches. And it also works for the situations that the object in the image small and there exsits multiple objects in the image. Interestingly, when keep ratio is very small such as 0.3, the left tokens not only fall onto the foreground area but also locate the distinguished parts of objects such as noses and ears of dogs, and the kept regions stays consistent with the highlight heatmap region of the base models. These demonstrate that mask fine-tuning mitigates inconsistencies between the base models used for initialization and dynamic vision transformers and enables stronger dynamic vision transformer.

\section{Conclusion}
Well-trained base models as initialization enables token pruning based dynamic vision transformers with faster convergence and better performance. But what is good initialization is not explored. We argue that there exist inconsistencies between currently used full training based models and token pruning based dynamic vision transformers, including calculation pattern, information amount and token selection strategies, leading to inferior performance. To tackle these inconsistencies, we introduce masked fine-tuning. Masked fine-tuning forces the model to predict the class labels of obscured images in information loss case, endowing base models with strong ability against occlusion and information loss, thus enabling stronger dynamic vision transformers. Extensive experiments on ImageNet show base models trained with masked fine-tuning get higher accuracy under masked input and token pruning settings. With our better initialization, different token pruning based dynamic vision transformers all achieve better performance. Moreover, our proposed masked fine-tuning achieves consistent improvements on supervised pre-trained models, constrastive learning based self-supervised pre-trained models, masked image modeling based self-supervised pre-trained models and models trained from scratch, which demonstrate its good generalization.

% Current token pruning based dynamic vision transformers usually base on plain transformers without hierarchical structure. Due to combination with dynamic vision transformers, our masked fine-tuning also lies on plain vision transformers and cannot apply into hierarchical transformers directly. Luckily, there are some works trying to realize masked image modeling on hierarchical transformer~\cite{huang2022green, li2022uniform}. We will leave adapting masked fine-tuning into hierarchical vision transformers in the future work.

\bibliographystyle{IEEEtran}
\bibliography{reference}
\end{document}